\documentclass[10pt,twocolumn,letterpaper]{article}

\usepackage{cvpr}              

\usepackage{times}
\usepackage{epsfig}
\usepackage{graphicx}
\usepackage{amsmath}
\usepackage{amssymb}
\usepackage{verbatim}
\usepackage{hhline}
\usepackage{multirow}
\usepackage{makecell}

\usepackage{xcolor,colortbl}

\usepackage{soul}

\usepackage{amssymb}
\usepackage{pifont}

\makeatletter
\DeclareRobustCommand\onedot{\futurelet\@let@token\@onedot}
\def\@onedot{\ifx\@let@token.\else.\null\fi\xspace}
\def\eg{\emph{e.g}\onedot} 
\def\ie{\emph{i.e}\onedot}

\def\etal{\emph{et al}\onedot}

\newcommand{\Tref}[1]{Table~\textcolor{blue}{\ref{#1}}}

\newcommand{\Fref}[1]{Fig.~\textcolor{blue}{\ref{#1}}}
\newcommand{\Sref}[1]{Sec.~\textcolor{blue}{\ref{#1}}}

\usepackage[accsupp]{axessibility}  

\usepackage{graphicx}
\usepackage{amsmath}
\usepackage{amssymb}
\usepackage{booktabs}
\usepackage{xspace}

\usepackage[pagebackref,breaklinks,colorlinks]{hyperref}

\usepackage[capitalize]{cleveref}
\crefname{section}{Sec.}{Secs.}
\Crefname{section}{Section}{Sections}
\Crefname{table}{Table}{Tables}
\crefname{table}{Tab.}{Tabs.}

\def\cvprPaperID{2684} 
\def\confName{CVPR}
\def\confYear{2022}

\begin{document}

\title{UDA-COPE: Unsupervised Domain Adaptation \\ for Category-level Object Pose Estimation}

\author{Taeyeop Lee\\
\and
Byeong-Uk Lee\\
\and
Inkyu Shin\\
\and
Jaesung Choe \\
\and
Ukcheol Shin\\
\and
In So Kweon \\
\and
Kuk-Jin Yoon \\
}
\affiliation{\!\!\!\!\!\!\!\!\!\!\!\!\!\!\!\!\!\!\!\!\!\!\!\!\!\!\!\!\!\!\!\!\!\!\!\!\!\!\!\!\!\!\!\!\!\!\!\!\!\!\!\!\!\!\!\!\!\!\!KAIST}


\maketitle

\begin{abstract}
Learning to estimate object pose often requires ground-truth (GT) labels, such as CAD model and absolute-scale object pose, which is expensive and laborious to obtain in the real world. To tackle this problem, we propose an unsupervised domain adaptation (UDA) for category-level object pose estimation, called \textbf{UDA-COPE}. Inspired by recent multi-modal UDA techniques, the proposed method exploits a teacher-student self-supervised learning scheme to train a pose estimation network without using target domain pose labels. We also introduce a bidirectional filtering method between the predicted normalized object coordinate space (NOCS) map and observed point cloud, to not only make our teacher network more robust to the target domain but also to provide more reliable pseudo labels for the student network training. Extensive experimental results demonstrate the effectiveness of our proposed method both quantitatively and qualitatively. Notably, without leveraging target-domain GT labels, our proposed method achieved comparable or sometimes superior performance to existing methods that depend on the GT labels.

\end{abstract}
\vspace{-5mm}



\section{Introduction}

Object pose estimation is one of the crucial tasks used in various robotics and computer vision applications for robot manipulation~\cite{zeng2017multi,wong2017segicp,wang2019densefusion,du2021vision} and augmented reality (AR)~\cite{runz2018maskfusion, marchand2015pose, marder2016project}.
Using sensor data such as images or point clouds, this task aims to estimate the poses of target objects including 3D orientation, 3D location, and size information.
 
Previous 6D object pose estimation methods follow the instance-level pose estimation schemes~\cite{tremblay2018corl:dope, xiang2017posecnn, park2019pix2pose, peng2019pvnet, wang2019densefusion, he2020pvn3d, he2021ffb6d} that rely on given 3D CAD model information (\eg, keypoints, geometry) and the size of known objects.
However, these methods typically have difficulty estimating the pose of unknown objects since they do not yet have 3D CAD models as priors.
%
%
%

%
In contrast to the instance-level scheme, category-level object pose estimation~\cite{wang2019normalized, Tian2020prior, chen2020cass, lin2021dualposenet, wang2021category, chen2021sgpa} approaches are more efficient in that a single network can infer multiple classes at once.
In particular, Wang~\etal~\cite{wang2019normalized} introduced a pioneering representation called Normalized Object Coordinate Space (NOCS), to align different object instances within one category in a shared 3D orientation.
By estimating per-category NOCS maps, it is able to estimate the 6D pose of unseen objects without prior 3D CAD models.
Its strengths have led to the use of NOCS representation in the following studies~\cite{Tian2020prior, chen2020cass, lin2021dualposenet, wang2021category, chen2021sgpa}. 

However, current object pose estimation research mostly relies on supervised learning, which requires expensive GT labels such as 3D object CAD models and absolute object pose.
These labels are not only difficult to obtain in the real world but are also unreliable due to the human-annotation.
Because of this difficulty, most of the training depends on synthetic datasets~\cite{sundermeyer2018implicit, kehl2017ssd, tremblay2018corl:dope} and is usually not feasible in real-world applications due to domain gaps.

To cope with the real-world data scarcity problem, we take a look at unsupervised domain adaptation (UDA) methods~\cite{jaritz2020xmuda, li2019bidirectional, zou2019confidence}.
UDA approaches often consider two types of datasets, the source domain (\ie synthetic dataset) and the target domain (\ie real-world dataset) dataset.
The main goal of the UDA methods is to successfully make deep learning networks robust to the target domain using only the GT labels of the source domain. Various techniques exist, such as pseudo label generation~\cite{jaritz2020xmuda, li2019bidirectional}, teacher and student networks with momentum updates~\cite{araslanov2021self, zhang2021prototypical}, adversarial learning~\cite{bousmalis2017unsupervised, lee2018diverse, bousmalis2018using}, and etc.

In this paper, we propose an Unsupervised Domain Adaptation for Category-level Object Pose Estimation (UDA-COPE).
The proposed method effectively transfers task knowledge from a synthetic domain to a real domain by exploiting a multi-modal self-supervised learning scheme using pseudo labels.
Our UDA-COPE concentrates on how to make high-quality pseudo-labels that are efficiently targeted for the category-level pose estimation task.
To this end, we designed {\emph{bidirectional point filtering}} to remove noisy and inaccurate points based on pose optimization.
Extensive experiments demonstrate that our UDA-COPE and bidirectional point filtering successfully can reduce the domain gap between synthetic and real datasets.
Moreover, our framework achieved better performance than the previous supervised methods~\cite{wang2019normalized, Tian2020prior, chen2020cass, wang2021category}.
The contributions of our method are summarized as:
\begin{itemize}
    \item We propose an RGB-D based Unsupervised Domain Adaptation for Category-level Object Pose Estimation (UDA-COPE) framework that addresses the problem of data deficiency in real-world scenarios. 
    \item We design a teacher-student framework where high-quality pose-aware pseudo labels can be obtained via the proposed bidirectional point filtering. 
    \item Our method shows comparable or sometimes better results than supervised pose estimation approaches.
\end{itemize}

\section{Related Work}

\noindent \textbf{Category-level object pose estimation.} \
This task~\cite{wang2019normalized, Tian2020prior, chen2020cass, chen2020category, lin2021dualposenet, chen2021sgpa} deals with objects of unseen instances but known categories.
%
The most recent category-level object pose and size methods~\cite{wang2019normalized, Tian2020prior, wang2021category, chen2021sgpa, lee2021category} use the dense Normalized Object Coordinate Space (NOCS) representation as a basic way to estimate pose. 
It aligns different object instances within one category in a shared 3D orientation. 
Shape Prior~\cite{Tian2020prior} improves the quality of NOCS maps by generating a representative shape prior and a deformation NOCS map for each category.
CR-Net~\cite{wang2021category} extends the shape-prior method~\cite{Tian2020prior} by using the cascade relation and recurrent reconstruction methods. Similarly, SGPA~\cite{chen2021sgpa} proposes a prior adaptation method. 

Despite notable improvements, these studies rely on fully supervised learning, and as such, they require a large amount of manually labeled data, such as the 6D pose of objects, 3D CAD models, and NOCS maps.
It is also time-consuming and expensive to create accurate GT information in real-world scenarios.
To the best of our knowledge, CPS++~\cite{manhardt2020cps++} is the only existing approach that addresses this data dependency problem.
It aims to predict 3D shape using an RGB image to optimize pose and utilizes an unsupervised learning scheme by computing consistency between the observed depth map and the rendered depth.
The rendered depth is obtained by projecting an estimated 3D shape with the predicted pose.
However, 3D shape reconstruction from a single image is challenging and makes their unsupervised guidance unreliable.
Recently, Li~\etal~\cite{li2021leveraging} suggested a self-supervised method that leverages SE(3) equivalent representation for category-level pose estimation. However, they do not fully consider pose parameters at the category-level that consist of 3D orientation, 3d location, and size information.


\noindent \textbf{Multi-modal UDA.} \
xMUDA~\cite{jaritz2020xmuda} is a pioneer approach for unsupervised domain adaptation (UDA) in 2D/3D semantic segmentation for multi-modal scenarios.
A few methods have considered an extra modality (\ie, depth) during training time and leveraged such privileged information to boost adaptation performance~\cite{lee2018diverse ,vu2019dada}.
Reza~\etal~\cite{loghmani2020unsupervised} proposed a multi-modal UDA in instance-level object pose estimation, but it only considered relative pose on a 2D image level. 
However, multi-modal UDA approaches have not yet been explored in the category-level object pose estimation task.

\section{Method}
\label{sec:method}
\begin{figure}
\begin{center}
\includegraphics[width=1.0\linewidth]{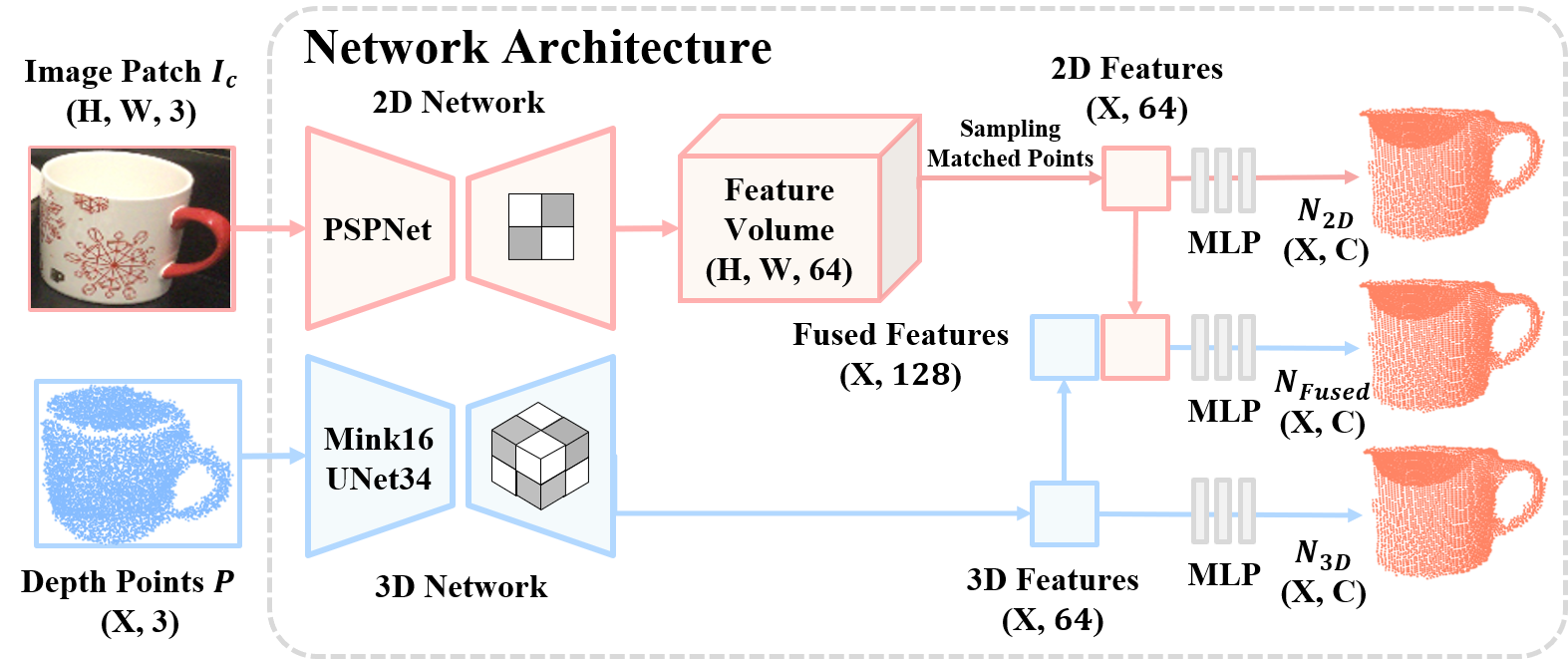}
\caption{\textbf{Network for multi-modal NOCS map estimation.}}
\vspace{-0.2in}
\label{fig:network}
\end{center}
\end{figure}


Given an RGB image $I$, point cloud $P$, and segmentation labels~$S$, our architecture aims to regress the 6D pose and size $s{\in}\mathbb{R}^3$ of objects. 
The 6D pose is defined as the rigid transformation of $[R|t]$: rotation $R{\in} SO(3)$, and translation $t{\in}\mathbb{R}^3$.
Following previous studies~\cite{Tian2020prior, chen2020cass, wang2021category, chen2021sgpa, lin2021dualposenet}, the segmentation labels~$S$ are used to crop the RGB images and point clouds.
We leverage the NOCS representation to align different object instances within one category in a shared orientation.
The categorical object pose $[R|t]$ and size $s$ are estimated by Umeyama algorithm~\cite{umeyama1991least} with RANSAC~\cite{fischler1981random}, which optimizes $[R|t]$ and $s$ by minimizing the distances between point cloud $P$ and an estimated NOCS map $N$.

We first illustrate our network architecture (\Sref{subsec:network_architecture}).
Then, we introduce training methods of supervised learning using synthetic dataset (\Sref{subsec:learning_scheme}) and unsupervised domain adaption using real-world dataset (\Sref{subsec:pose_aware_pseudo_labels}).


\begin{figure*}
\begin{center}
\includegraphics[width=0.95\linewidth]{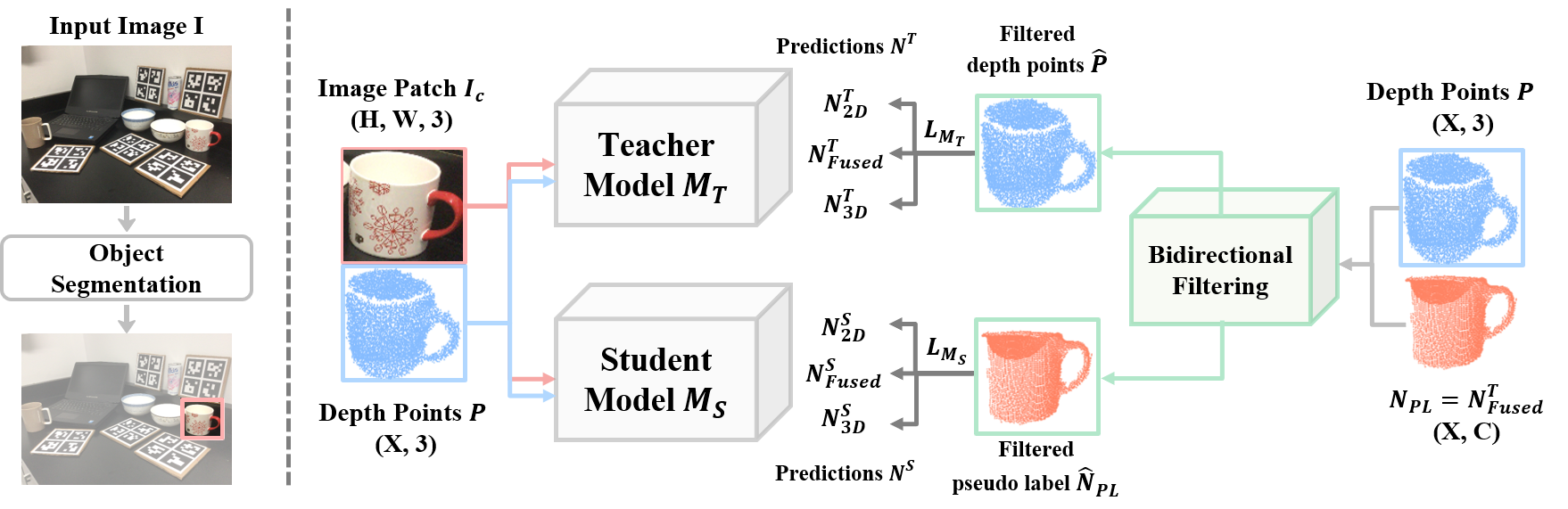}
\vspace{-1.0mm}
\caption{\textbf{Overview of unsupervised domain adaptation for category-level object pose estimation (UDA-COPE).} 
UDA-COPE utilizes pseudo label based teacher/student training scheme. Our proposed bidirectional point filtering method removes the noisy pseudo labels and gives reliable guidance to the student network. At the same time, filtered depth points gives additional self-supervision to the teacher network so that it can be robust to the domain gap between the synthetic and real dataset.
}
\vspace{-0.2in}
\label{fig:uda_cope}
\end{center}
\end{figure*}

\subsection{Network Architecture}
\label{subsec:network_architecture}
Recent category-level object pose estimation methods~\cite{Tian2020prior, chen2020cass, lin2021dualposenet} take an RGB-D input to extract the 2D/3D features. 
We designed separate 2D/3D branches to extract features from both modalities.
We use PSPNet~\cite{zhao2017pyramid} with ResNet34~\cite{he2016deep} for 2D feature extraction and the Mink16UNet34~\cite{choy20194d} for 3D feature extraction.
At this time, the 2D feature is extracted by sampling features that are validly matched with point cloud $P$ from the feature volume. 
Finally, we have a fused branch that combines each feature from both branches.
Every branch estimates a NOCS map~($N$) with a separate NOCS header, which consists of three multi-layer perceptrons (MLP) layers.
Our multi-modal NOCS map prediction network is illustrated in \Fref{fig:network}.
We designate the NOCS map estimation of each branch as $N_{\text{2D}}$, $N_{\text{3D}}$, $N_{\text{Fused}}$, according to respective feature property.


\subsection{Pre-Training with Synthetic Data}
\label{subsec:learning_scheme}
Inspired by pseudo label (PL) based methods~\cite{jaritz2020xmuda, li2019bidirectional}, our method consists of a teacher and a student model.
\Fref{fig:uda_cope} shows the overview of our teacher and student model.
The initial prediction of teacher model $M_\text{T}$ becomes a pseudo label $N_{\text{PL}}$ for a student model, and student model $M_\text{S}$ learns from the pseudo label as a GT.
Our teacher and student model have the same structure as was described in~\Sref{subsec:network_architecture}.

We first train our teacher model in a supervised manner using the labeled synthetic dataset.
For the NOCS map prediction using the GT information, we utilize cross-entropy loss, as in, $H(N_\text{gt}, N^\text{T})$, where the supervision is given to all predictions from three branches.
Additionally, to make our teacher network more robust, we apply the 2D image and 3D points augmentation and use consistency loss $L_\text{C}$ so that each modality can output consistent results.
Total loss for the teacher network on the synthetic dataset is formulated as follows: 
\begin{equation}
\label{eq:teacher_training}
\begin{split}
    L_{M_\text{T}} &= \lambda_\text{N}H(N_\text{gt}, N^\text{T}) + \lambda_\text{C}L_\text{C}, \\
    L_\text{C} &= H(N^\text{T}, N^\text{T}_\text{Aug})
\end{split}
\end{equation}
where $N^\text{T}_\text{Aug}$ is the NOCS map prediction from the augmented input, and $\lambda_\text{N}$ and $\lambda_\text{C}$ are weighting parameters.
Notation for the modality of the predictions is discarded for better readability.

\begin{figure}
\begin{center}
\footnotesize
\begin{tabular}{c@{\hskip 0.005\linewidth}c@{\hskip 0.005\linewidth}c@{\hskip 0.005\linewidth}c}
\includegraphics[width=1.0\linewidth]{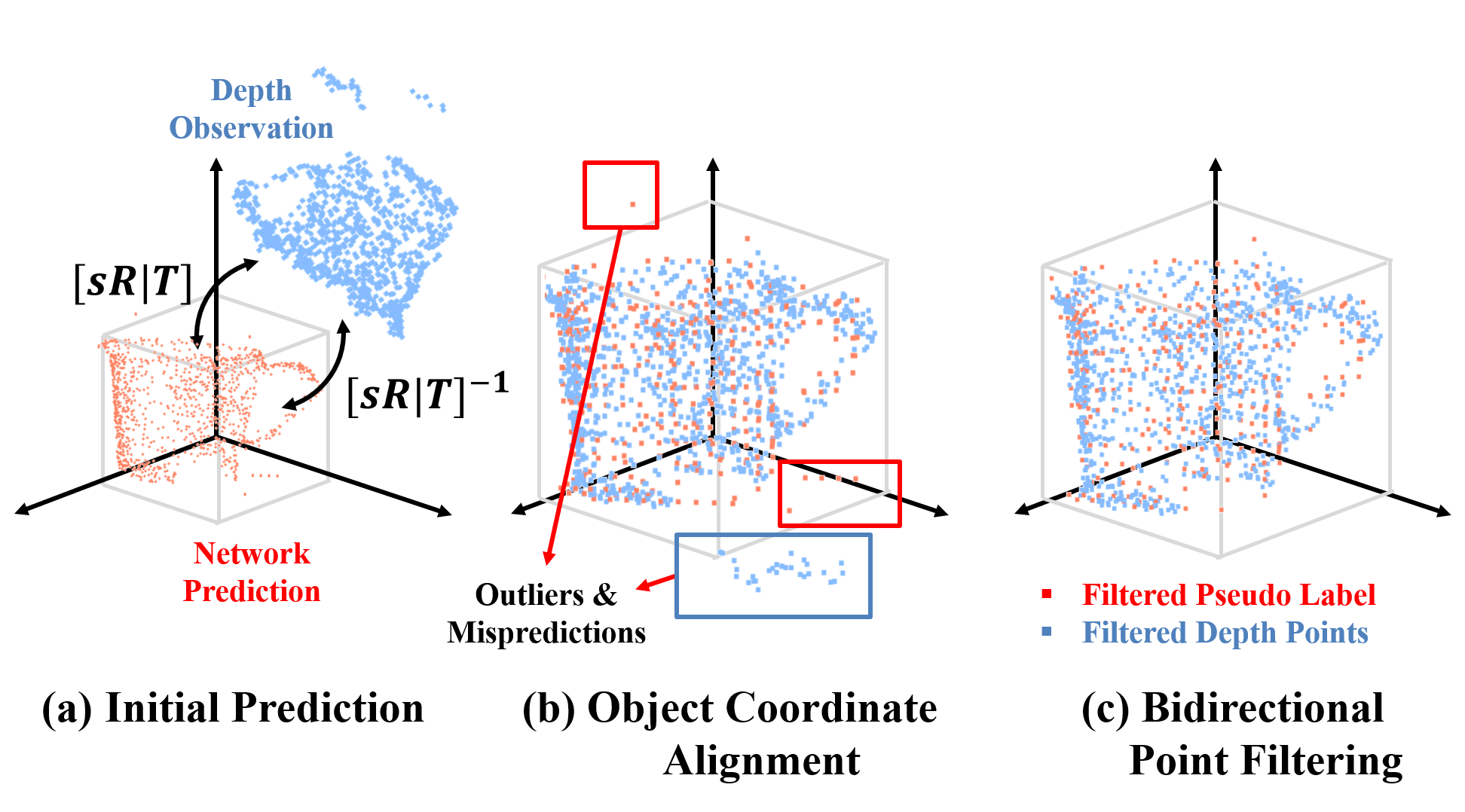}
 \\
\end{tabular}
\end{center}
\vspace{-1.0mm}
\vspace{-0.2in}
\caption{\textbf{Overview of bidirectional point filtering method.} Given pseudo labels and depth points (a), we estimate the pose and size using the Umeyama~\cite{umeyama1991least} algorithm and RANSAC~\cite{fischler1981random}, and align the depth points to normalized object coordinate (b). The pseudo label (red) and aligned depth points (blue) have noisy and inaccurate points. After our bidirectional point filtering, the noisy points are removed to give more reliable supervision for both teacher and student (c).}
\vspace{-0.2in}
\label{fig:bidirection}
\end{figure}


\subsection{Pose-Aware Unsupervised Domain Adaptation}
\label{subsec:pose_aware_pseudo_labels}
After training from the synthetic dataset, the most straight-forward yet naive approach is to train the student network using the prediction of the teacher network.
However, using the initial prediction from the teacher model as a pseudo label can be risky.
The risk is due to the lack of robustness of the teacher model itself, or more importantly, because of the insufficient knowledge that the teacher model holds with respect to real-world scenarios, due to the domain gap between the simulated and real worlds.
Techniques such as data augmentation and momentum update might help the feasibility but are still restricted.
Therefore, we need additional guidance for the teacher model to estimate high-quality predictions, and more reliable pseudo labels for our student model to learn from.

\subsubsection{Bidirectional Point Filtering}
\label{subsec:bidirection}
To solve these problems, we propose the bidirectional point filtering method which simultaneously removes the noise of the pseudo labels for the student and filters noisy depth points $P$ for a teacher network.
\Fref{fig:bidirection} shows an overview of the proposed bidirectional filtering method.
Our bidirectional filtering method uses the $P$ and $N_\text{PL}$ as input and initially estimates the pose $[R|t]$ and size $s$ using the Umeyama algorithm~\cite{umeyama1991least} with RANSAC~\cite{fischler1981random}.
Then it aligns the depth points $P$ to the NOCS coordinate by applying the inverse of the estimated pose, as in multiplying the matrix $[sR|t]^{-1}$.
We denote aligned depth points as $P'$.
And then we calculate the point-wise 3D distance $d$ between the aligned depth points $P'$ and pseudo label $N_\text{PL}$ to filter out noisy points from both sides using $\rho$ as the threshold.
Finally, we get the refined pseudo label $\hat{N}_\text{PL}$ and filtered aligned depth $\hat{P}$.
Our bidirectional point filtering can be expressed as:
\begin{equation}
\begin{split}
d(n) &= \| P'(n) - N_\text{PL}(n) \|~~\text{where}~~\forall n \in [1 |P'], \\
\hat{N}_\text{PL} &= \{ N_\text{PL}(n) : d(n) < \rho \}, \\
\hat{P} &= \{ P'(n)~~: d(n) < \rho \}, \\
\end{split}
\end{equation}

\Fref{fig:bidirection} shows that our bidirectional filtering method removes outliers of pseudo label $N_\text{PL}$ and depth $P$, and results in refined pseudo label $\hat{N}_\text{PL}$ and filtered depth points $\hat{P}$. 

\subsubsection{Self-Supervised Learning}
After the bidirectional filtering, we jointly train the teacher network and student network using the filtered pseudo labels $\hat{N}_\text{PL}$ and filtered aligned depth points $\hat{P}$.
Noted that we only use the filtered points $\hat{P}$ for the teacher training, which may be a smaller subset of an original $P$.
We use cross-entropy loss to train a student model using clean pseudo labels $\hat{N}_\text{PL}$.
The student model loss is defined as:
\begin{equation}
    L_{M_\text{S}} = - \frac{1}{|\hat{N_\text{PL}}|}\sum_{n=1}^{|\hat{N}_\text{PL}|} H(\hat{N}_\text{PL}(n), N^\text{S}(n)), 
\label{filtering_pseudo_label}
\end{equation}
where ${N}^{S}$ is the predictions of our student network.
At the same time, the teacher learns real data knowledge from observation.
We use cross-entropy loss by utilizing geometric consistency between our filtered aligned depth $ \hat{P}$ and estimated pseudo labels $N^T$.
The teacher model loss is defined as:
\begin{equation}
    L_{M_\text{T}} = - \frac{1}{|\hat{P}|}\sum_{n=1}^{|\hat{P}|} H(\hat{P}(n), N^\text{T}(n)).
\label{self_supervised_learning}
\end{equation}
We train our teacher model with a small learning rate for stable teacher network training.
For both teacher and student models, we compute the loss for all estimations, $N_\text{2D}$, $N_\text{3D}$, $N_\text{Fused}$, which shows better result than applying the loss to only $N_\text{Fused}$. 
We denote all estimation losses as all modality (AM) loss.


\section{Experiments}

\noindent{\textbf{Datasets.}}
We employ two commonly used category-level pose estimation datasets, a synthetic dataset and a real dataset. 
The synthetic dataset (source domain) is the Context-Aware MixedEd ReAlity (CAMERA) dataset \cite{wang2019normalized}, generated by rendering and compositing synthetic objects into real images in a context-aware manner.
It comprises 275k synthetic images for training.
The dataset contains 1085 object instances selected from 6 different categories - bottle, bowl, camera, can, laptop and mug.
We use the REAL dataset \cite{wang2019normalized} as a real dataset (target domain).
Compared to previous methods~\cite{wang2019normalized, Tian2020prior, chen2020cass, chen2021sgpa}, our method trains without object pose, 3D CAD, and NOCS map GT labels of the target domain.
It consists of 43,000 real world images of 7 scenes for training and 2,750 real world images of 6 scenes for evaluation.
The REAL evaluation set was designated REAL275.
We evaluated our method on a standard REAL275 benchmark for the task of category-level object pose estimation.

\noindent{\textbf{Metrics.}}
We followed the previous pose and size evaluation metric from Wang~\etal~\cite{wang2019normalized}, which evaluated the performance of 3D object detection and 6D pose estimation.
We report the average precision at different Intersection-Over-Union (IoU) thresholds for 3D object detection.
Threshold values of 25\%, 50\%, and 75\% were used to evaluate the results.
For 6D object pose evaluation, the average precision was computed considering rotation and translation errors.
For example, the $10^{\circ}, 10$cm metric denotes the percentage of object instances where the prediction error was less than $10^{\circ}$ and $10$cm.

\begin{table*}
\centering
    \resizebox{0.92\linewidth}{!}{
\begin{tabular}{c c ccc cc cccc}
\Xhline{4\arrayrulewidth}
\multirow{2}{*}{Method} & \multirow{2}{*}{Input} & \multirow{2}{*}{Syn} & \multirow{2}{*}{\begin{tabular}[c]{@{}c@{}}Real \\ w/ Label\end{tabular}} & \multirow{2}{*}{\begin{tabular}[c]{@{}c@{}}Real \\ w/o Label\end{tabular}} & \multicolumn{6}{c}{mAP (↑)}                                       \\ 
                        &                        &                      &                                                                                  &                                                                                   & $3D_{50}$ & $3D_{75}$ & $5^{\circ} 2cm$ & $5^{\circ} 5cm$ & $10^{\circ} 2cm$ & $10^{\circ} 5cm$ \\ \Xhline{2\arrayrulewidth}
CPS++~\cite{manhardt2020cps++}                   & RGB                     & \checkmark            &                                                                                  &                                                                                   & \textbf{72.6}        & -           & -      & \textbf{25.8}    & -       & -        \\
Metric Scale~\cite{lee2021category}                   & RGB                     & \checkmark            &                                                                                  &                                                                                   & 68.1        & 32.9           & 2.2      & 5.3    & 10.0       & 24.7        \\
NOCS~\cite{wang2019normalized}                    & RGB                     & \checkmark            &                                                                                  &                                                                                   & 36.7        & 3.4         & -      & 3.4     & -       & 20.4     \\
SPD~\cite{Tian2020prior}             & RGB-D                   & \checkmark            &                                                                                  &                                                                                   & 71.0        & \textbf{43.1}        & \textbf{11.4}   & 12.0    & \textbf{33.5}    & \textbf{37.8}     \\ \hline
NOCS~\cite{wang2019normalized}                    & RGB                     & \checkmark            & \checkmark                                                                        &                                                                                   & 78.0        & 30.1        & 7.2    & 10.0    & 13.8    & 25.2     \\
SPD~\cite{Tian2020prior}             & RGB                     & \checkmark            & \checkmark                                                                        &                                                                                   & 75.2        & 46.5        & 15.7   & 18.8    & 33.7    & 47.4     \\
SPD~\cite{Tian2020prior}             & RGB-D                   & \checkmark            & \checkmark                                                                        &                                                                                   & 77.4        & 53.5        & 19.5   & 21.6    & 43.5    & 54.0     \\
CASS~\cite{chen2020cass}                    & RGB-D                   & \checkmark            & \checkmark                                                                        &                                                                                   & 77.7        & -           & -      & 23.5    & -       & 58.0     \\
CR-Net~\cite{wang2021category}                  & RGB-D                  & \checkmark            & \checkmark                                                                        &                                                                                   & 79.3        & 55.9        & 27.8   & 34.3    & 47.2    & 60.8     \\
DualPoseNet~\cite{lin2021dualposenet}             & RGB-D                   & \checkmark            & \checkmark                                                                        &                                                                                   & 79.8        & \textbf{62.2}        & 29.3   & 35.9    & 50.0    & 66.8     \\
SGPA~\cite{chen2021sgpa}                    & RGB-D                   & \checkmark            & \checkmark                                                                        &                                                                                   & \textbf{80.1}        & 61.9        & \textbf{35.9}   & \textbf{39.6}    & \textbf{61.3}    & \textbf{70.7}     \\ \hline
CPS++~\cite{manhardt2020cps++}                   & RGB                     & \checkmark            &                                                                                  & \checkmark                                                                         & 72.8        & -           & -      & 25.2    & -       & -        \\
Ours                    & RGB                     & \checkmark            &                                                                                  & \checkmark                                                                         & 82.0        & 59.0        & 24.4   & 27.0    & 49.3    & 54.8     \\
Ours                    & D                   & \checkmark            &                                                                                  & \checkmark                                                                         & 79.6        & 57.8        & 21.2   & 29.1    & 48.7    & 65.9     \\
Ours                    & RGB-D                  & \checkmark            &                                                                                  & \checkmark                                                                         & \textbf{82.6}        & \textbf{62.5}        & \textbf{30.4}   & \textbf{34.8}    & \textbf{56.9}    & \textbf{66.0}     \\ \Xhline{4\arrayrulewidth}
\end{tabular}
}
\caption{\textbf{Quantitative comparison with state-of-the art methods on the REAL275 dataset.} Empty entries either could not be evaluated or were not reported in the original paper. }
    \label{tab:sota_table_all}
\end{table*}

\begin{table*}
\centering
    \resizebox{1.0\linewidth}{!}{
\begin{tabular}{c cc ccc ccccc}
\Xhline{4\arrayrulewidth}
\multirow{2}{*}{Method} & \multirow{2}{*}{Syn} & \multirow{2}{*}{\begin{tabular}[c]{@{}c@{}}Real \\ w/o Label\end{tabular}} & \multicolumn{8}{c}{mAP (↑)}                                                                 \\ 
                        &                      &                                                                                   & $3D_{25}$ & $3D_{50}$ & $3D_{75}$ & $5^{\circ} 2cm$ & $5^{\circ} 5cm$ & $10^{\circ} 2cm$ & $10^{\circ} 5cm$ & $10^{\circ} 10cm$ \\ \Xhline{2\arrayrulewidth}
CPS++ (RGB)             & \checkmark            &                                                                                   & 84.5        & 72.6        & -           & -      & 25.8    & -       & -        & 55.4      \\
CPS++ (RGB)             & \checkmark            & \checkmark                                                                         & 84.6 \textcolor{blue}{\footnotesize{(+0.1)}}        & 72.8 \textcolor{blue}{\footnotesize{(+0.2)}}        & -           & -      & 25.2 \textcolor{red}{\footnotesize{(-0.6)}}    & -       & -        & 58.6 \textcolor{blue}{\footnotesize{(+3.2)}}      \\ \hline
Ours (RGB)              & \checkmark            &                                                                                   & 83.3        & 79.9        & 49.7        & 15.4   & 18.3    & 37.6    & 46.7     & 48.9      \\
Ours (RGB)              & \checkmark            & \checkmark                                                                         & 83.8 \textcolor{blue}{\footnotesize{(+0.5)}}         & 82.0 \textcolor{blue}{\footnotesize{(+2.1)}}        & 59.0 \textcolor{blue}{\footnotesize{(+9.3)}}        & 24.4 \textcolor{blue}{\footnotesize{(+9.0)}}   & 27.0 \textcolor{blue}{\footnotesize{(+8.7)}}    & 49.3 \textcolor{blue}{\footnotesize{(+11.7)}}    & 54.8 \textcolor{blue}{\footnotesize{(+8.1)}}     & 56.9 \textcolor{blue}{\footnotesize{(+8.0)}}      \\ 
Ours (RGB-D)              & \checkmark            & \checkmark                                                                         & 84.0 \textcolor{blue}{\footnotesize{(+0.7)}}        & 82.6 \textcolor{blue}{\footnotesize{(+2.7)}}        & 62.5 \textcolor{blue}{\footnotesize{(+12.8)}}        & 30.4 \textcolor{blue}{\footnotesize{(+15.0)}}   & 34.8 \textcolor{blue}{\footnotesize{(+16.5)}}    & 56.9 \textcolor{blue}{\footnotesize{(+19.3)}}    & 66.0 \textcolor{blue}{\footnotesize{(+19.3)}}     & 68.3 \textcolor{blue}{\footnotesize{(+19.4)}}      \\ \Xhline{4\arrayrulewidth}
\end{tabular}
}
\caption{\textbf{Quantitative comparison of unsupervised pose estimation approaches on the REAL275 dataset.} Empty entries are either not able to be evaluated or not reported in the original paper. Performance margins are calculated compared to the synthetic-only results.}
\label{tab:rgb_only_comparison}
\vspace{-0.2in}
\end{table*}

\subsection{Implementation Details}
\label{subsec:implementation_details} 
We implemented our method with PyTorch~\cite{paszke2019pytorch}.
To obtain object regions from the real-world dataset, we used the GT segmentation labels during training and the off-the-shelf object segmentation method, Mask R-CNN~\cite{he2017mask}, during inference.
For a detected instance, we resized the image patch to 192 x 192, and randomly sampled 1024 point clouds.
The 2D and 3D features had 64 dimensions and the fused feature had 128 dimension features, due to concatenation.
We trained our teacher network for 50 epochs on the synthetic dataset.
We used the Adam optimizer with an initial learning rate of 0.0001 and a batch size of 16, where the learning rate was decreased by a factor of 0.6, 0.3, 0.1, 0.01 at 15k, 30k, 45k, 60k iterations, respectively.
We set $\lambda_\text{N}$ = 1.0, $\lambda_\text{C}$ = $1e{-}6$ for the teacher network.
The $\rho$ was set to 0.05 for our proposed bidirectional filtering.
On the unlabeled real dataset, both the teacher and student networks were trained for 1 epoch.
The learning parameters for our student network were the same as those for the teacher training scheme on the synthetic dataset.
We set the learning rate for real-world self-supervised learning for the teacher network to be $1e{-}7$. 

\subsection{Comparison with State-of-the-art}
\label{subsec:sota_comparison}
We compared our methods with state-of-the-art methods that were trained on different datasets and labels: 1) labeled synthetic dataset, 2) labeled synthetic and real datasets, 3) labeled synthetic and unlabeled real datasets.
All methods were evaluated on the REAL275 dataset.
Note that only the approaches with the ability to perform multi-class category-level pose estimation using a single network were considered.
RGB, Depth and RGB-D denotes the modality of the network input, and most of the RGB based approaches utilize depth information in the pose optimization or refinement process.

\begin{table*}
\centering
    \resizebox{1.0\linewidth}{!}{
\def\arraystretch{1.3}
\begin{tabular}{l cccc c cccc c cccc}
\Xhline{4\arrayrulewidth}
\multicolumn{1}{c}{\multirow{2}{*}{Method}} 
& \multicolumn{4}{c}{RGB}                                          && \multicolumn{4}{c}{Depth}                                        && \multicolumn{4}{c}{RGB-D}                    \\ \cline{2-5} \cline{7-10}  \cline{12-15} 
\multicolumn{1}{c}{}                        & $3D_{50}$ & $3D_{75}$ & $5^{\circ} 2cm$& \multicolumn{1}{c}{$5^{\circ} 5cm$} && $3D_{50}$ & $3D_{75}$ & $5^{\circ} 2cm$ & \multicolumn{1}{c}{$5^{\circ} 5cm$} && $3D_{50}$ & $3D_{75}$ & $5^{\circ} 2cm$ & $5^{\circ} 5cm$ \\ \Xhline{2\arrayrulewidth}
Lower Bound                                  & 79.9        & 49.7        & 15.4   & \multicolumn{1}{c}{18.3}    && 76.7        & 52.7        & 14.9   & \multicolumn{1}{c}{22.7}    && 80.7        & 60.9        & 23.0   & 27.9    \\ \hline
PL                                           & 78.7        & 51.0        & 11.6   & \multicolumn{1}{c}{13.5}    && 76.2        & 51.0        & 15.4   & \multicolumn{1}{c}{23.0}    && 80.3        & 58.6        & 23.8   & 28.2    \\
PL + MU                                      & 79.0        & 52.8        & 11.9   & \multicolumn{1}{c}{13.6}    && 76.9        & 51.3        & 14.7   & \multicolumn{1}{c}{22.7}    && 80.0        & 58.3        & 23.3   & 27.8    \\
PL + AM                                      & 80.2        & 56.2        & 21.2   & \multicolumn{1}{c}{24.8}    && 77.1        & 57.9        & 18.2   & \multicolumn{1}{c}{24.7}    && 81.6        & 60.8        & 24.6   & 29.2    \\
PL + xMUDA                                           & 78.4        & 55.3        & 21.9   & \multicolumn{1}{c}{25.5}    && 76.5        & 56.6        & 18.3   & \multicolumn{1}{c}{26.0}    && 80.7        & 60.2        & 25.7   & 30.3    \\
PL-F + AM                                    & 81.5        & 59.0        & 23.3   & \multicolumn{1}{c}{26.1}    && 77.3        & \textbf{58.4}        & 20.0   & \multicolumn{1}{c}{27.3}    && 81.3        & 62.0        & 28.1   & 32.9    \\
PL-F + AM + TSL                              & \textbf{82.0} \textcolor{blue}{\footnotesize{(+2.1)}}        & \textbf{59.0} \textcolor{blue}{\footnotesize{(+9.3)}}       & \textbf{24.4} \textcolor{blue}{\footnotesize{(+9.0)}}   & \multicolumn{1}{c}{\textbf{27.0} \textcolor{blue}{\footnotesize{(+8.7)}}}    && \textbf{79.6} \textcolor{blue}{\footnotesize{(+2.9)}}        & 57.8 \textcolor{blue}{\footnotesize{(+5.1)}}        & \textbf{21.2} \textcolor{blue}{\footnotesize{(+6.3)}}   & \multicolumn{1}{c}{\textbf{29.1} \textcolor{blue}{\footnotesize{(+6.4)}}}    && \textbf{82.6} \textcolor{blue}{\footnotesize{(+1.9)}}        & \textbf{62.5} \textcolor{blue}{\footnotesize{(+1.6)}}        & \textbf{30.4} \textcolor{blue}{\footnotesize{(+7.4)}}   & \textbf{34.8} \textcolor{blue}{\footnotesize{(+6.9)}}    \\ \hline
Upper Bound                                  & 82.7        & 66.7        & 29.3   & \multicolumn{1}{c}{32.8}    && 79.9        & 64.9        & 23.7   & \multicolumn{1}{c}{29.6}    && 82.9        & 70.4        & 31.8   & 35.8    \\ \Xhline{4\arrayrulewidth}
\end{tabular}
}
\caption{\textbf{Ablation studies on UDA components.} Lower Bound: trained with labeled source only, Upper Bound: trained with both labeled source and target, PL: Pseudo Label, MU: Momentum Update, AM: All Modality loss, PL-F: Pseudo Label Filtering, TSL: Teacher Self-supervised Learning~\eqref{self_supervised_learning}. Performance margins are calculated compared to the Lower Bound.}
    \label{tab:ablation_study_all}
\end{table*}

\noindent{\textbf{Supervised Pose Estimation methods.}}
\Tref{tab:sota_table_all} summarizes the results of the state-of-the-art category-level object pose estimation methods.
Obviously, supervised training with the real data annotation significantly improved the overall performance, as are revealed by comparing the results of NOCS~\cite{wang2019normalized} and SPD~\cite{Tian2020prior} on different training dataset conditions.
However, our unsupervised method showed results superior to NOCS~\cite{wang2019normalized}, SPD~\cite{Tian2020prior}, CASS~\cite{chen2020cass}, and CR-Net~\cite{wang2021category}.
Compared to two of the strongest previous approaches, SGPA~\cite{chen2021sgpa} and DualPoseNet~\cite{lin2021dualposenet}, ours still showed comparable performance.
This indicates that our proposed filtered pseudo label based UDA-COPE is robust when estimating object pose in unseen real-world instances.

\noindent{\textbf{Unsupervised Pose Estimation methods.}} 
\Tref{tab:rgb_only_comparison} summarizes the results of CPS++ and our method on source only, and source with unlabeled target training conditions.
CPS++~\cite{manhardt2020cps++} provides self-supervision by computing the consistency between the observed depth map and the rendered depth.
The rendered depth is obtained by projecting an estimated 3D shape with the predicted pose.
The results from row 1 and row 2 in \Tref{tab:rgb_only_comparison} show that for CPS++, using unlabeled real data marginally improved performance, and sometimes even worsened it, as in $5^{\circ}, 5$cm metric.
We believe that their self-supervision is unreliable because of ambiguous 3D shape reconstruction using only a single-view RGB image.

Comparing row 3 and row 4, it can be seen that our proposed method shows improved results for every metrics, with some metrics showing notable margins such as an 8.7 mAP (48\%) increase in $5^{\circ}, 5$cm.
Also, in the last row, our RGB-D result had better performance than the single modality based outputs.
Therefore, we claim that our proposed algorithm is more effective by utilizing a pseudo label based learning scheme with modality and pose-aware self-supervision. 
The effectiveness of each components will be ablated thoroughly in the following sections.

\subsection{Ablation Study}
\label{subsec:ablation_study}
Our ablation studies were conducted using the predictions from all branches: RGB, Depth, and RGB-D.
The lower bound and upper bound in \Tref{tab:ablation_study_all} are the results from using a single network described in \Sref{subsec:network_architecture}.
The lower bound was trained source only, while the upper bound utilized both source and target data with their labels.
\Tref{tab:ablation_study_all} summarizes the results of the ablation studies.

\begin{table*}
\centering
    \resizebox{1.0\linewidth}{!}{
\def\arraystretch{1.3}
\begin{tabular}{c cccc c cccc c cccc}
\Xhline{4\arrayrulewidth}
\multirow{2}{*}{Filter}  & \multicolumn{4}{c}{RGB}                                          && \multicolumn{4}{c}{Depth}                                        && \multicolumn{4}{c}{RGB-D}                    \\ \cline{2-5} \cline{7-10} \cline{12-15}
                        & $3D_{50}$ & $3D_{75}$ & $5^{\circ} 2cm$ & \multicolumn{1}{c}{$5^{\circ} 5cm$} && $3D_{50}$ & $3D_{75}$ & $5^{\circ} 2cm$& \multicolumn{1}{c}{$5^{\circ} 5cm$} && $3D_{50}$ & $3D_{75}$ & $5^{\circ} 2cm$ & $5^{\circ} 5cm$ \\ \Xhline{2\arrayrulewidth}
None                    & 80.2        & 56.2        & 21.2   & \multicolumn{1}{c}{24.8}    && 77.1        & 57.9        & 18.2   & \multicolumn{1}{c}{24.7}    && 81.6        & 60.8        & 24.6   & 29.2    \\ \hline
Top k (conf)             & 79.0        & 55.5        & 21.6   & \multicolumn{1}{c}{25.0}    && 76.7        & 56.4        & 18.1   & \multicolumn{1}{c}{24.8}    && 81.4        & 60.4        & 25.6   & 30.1    \\
Top k (conf, class-wise)            & 78.8        & 55.0        & 20.4   & \multicolumn{1}{c}{23.8}    && 76.7        & 56.2        & 18.1   & \multicolumn{1}{c}{24.9}    && 80.9        & 59.7        & 25.3   & 29.8    \\
Entropy                 & 77.6        & 54.9        & \textbf{24.1}   & \multicolumn{1}{c}{\textbf{27.7}}    && \textbf{77.7}       & 57.0        & 18.4   & \multicolumn{1}{c}{25.3}    && 80.6        & 60.8        & 26.6   & 31.1    \\
SoftMax Max                & 79.9        & 52.7        & 19.6   & \multicolumn{1}{c}{22.4}    && 76.9        & 56.3        & 17.4   & \multicolumn{1}{c}{25.5}    && 79.9        & 59.2        & 25.6   & 30.0    \\
ArgMax Match                 & 79.6        & 52.2        & 17.7   & \multicolumn{1}{c}{19.9}    && 76.2        & 56.0        & \textbf{20.1}   & \multicolumn{1}{c}{\textbf{27.6}}    && \textbf{81.6}        & 60.9        & 24.1   & 27.5    \\
Softmax Avg.            & 79.1        & 55.9        & 21.8   & \multicolumn{1}{c}{25.3}    && 77.1        & 55.9        & 18.6   & \multicolumn{1}{c}{26.1}    && 81.3        & 61.9        & 25.5   & 30.0    \\ \Xhline{2\arrayrulewidth}
Ours                    & \textbf{81.5} \textcolor{blue}{\footnotesize{(+1.3)}}        & \textbf{59.0} \textcolor{blue}{\footnotesize{(+2.8)}}        & 23.3 \textcolor{blue}{\footnotesize{(+2.1)}}   & \multicolumn{1}{c}{26.1 \textcolor{blue}{\footnotesize{(+1.3)}}}    && 77.3 \textcolor{blue}{\footnotesize{(+0.2)}}        & \textbf{58.4} \textcolor{blue}{\footnotesize{(+0.5)}}        & 20.0 \textcolor{blue}{\footnotesize{(+1.8)}}   & \multicolumn{1}{c}{27.3 \textcolor{blue}{\footnotesize{(+2.6)}}}    && 81.3 \textcolor{red}{\footnotesize{(-0.3)}}        & \textbf{62.0} \textcolor{blue}{\footnotesize{(+1.2)}}        & \textbf{28.1} \textcolor{blue}{\footnotesize{(+3.5)}}   & \textbf{32.9} \textcolor{blue}{\footnotesize{(+3.7)}}    \\ \Xhline{4\arrayrulewidth}

\end{tabular}
}
\caption{\textbf{Ablation study of the pseudo label filtering methods.} Performance margins were calculated as compared to the results without using any pseudo label filtering.}
    \label{tab:filter_rgb}
\end{table*}

\begin{figure*}
\begin{center}
\includegraphics[width=1.0\linewidth]{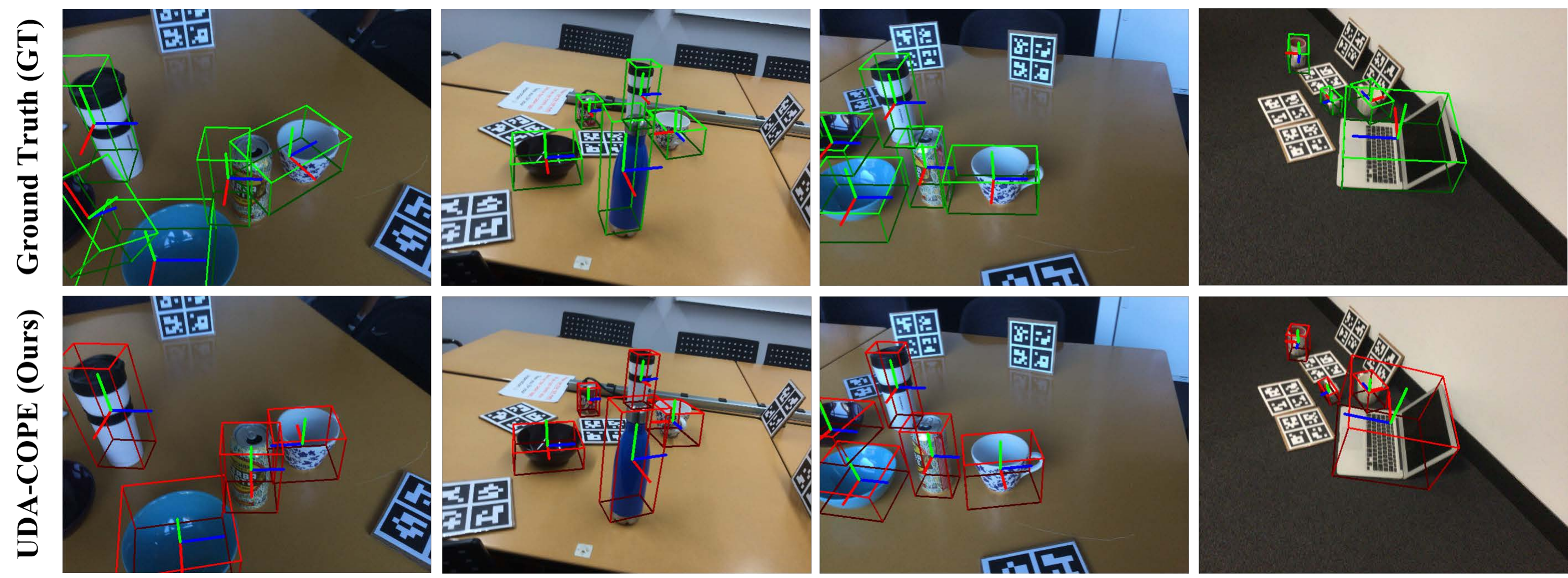}
\vspace{-0.1in}
\caption{\textbf{Noisy GT label examples of the Real training dataset.} Human-annotated GT pose labels on the real dataset (top row) are sometimes more inaccurate than our predicted pseudo labels (bottom row).}
\label{fig:gt_error}
\end{center}
\vspace{-0.1in}
\end{figure*}

\noindent{\textbf{Naive Teacher/Student UDA for Pose Estimation.}}
To verify the effectiveness of our proposed pose-aware UDA methods, we first applied a very naive version of teacher-student learning scheme with the momentum update.
When applying the momentum update, we updated our teacher network every 100 iterations with a $\gamma$ of 0.999.

However, it can be seen that neither methods were not very helpful.
The possible reasons are: 1) directly using the initial predictions of our teacher network as a pseudo label was unreliable due to the domain gap and noisy prediction, 2) providing supervision only to the fused branch may restrict effective training of the 2D and 3D branches, which disrupt the fused branch after all, and 3) the momentum update using the gradient information retrieved from the unreliable pseudo label and thereby unreliable student network training were insufficient for the teacher network to learn real world adaptation.

\noindent{\textbf{Modality-Aware UDA methods.}}
We applied all modality (AM) loss which computes cross-entropy loss between a pseudo label and all outputs from the three branches.
Comparing (PL) and (PL + AM) in \Tref{tab:ablation_study_all}, shows that the performances of the RGB-based prediction and Depth-based prediction improved by large margins. Accordingly, we achieved improved accuracy from the RGB-D-based fused branch as well.

To compare our methods with the previous multi-modal unsupervised domain adaptation method, we also applied xMUDA~\cite{jaritz2020xmuda} which constraints the 2D feature and 3D feature for consistency.
xMUDA consistency was given during the student network training.
xMUDA showed a similar amount of performance boost as our AM loss.
This is because both AM and xMUDA help our network to output consistent predictions between $N_\text{2D}$, $N_\text{3D}$, and $N_\text{Fused}$.
We chose AM since it can provide more straight-forward supervision to each branch.

\noindent{\textbf{Pose-Aware UDA methods.}}
We compared our two major components, which were specifically designed for pose estimation.
As shown in \Tref{tab:ablation_study_all}, pseudo label filtering (PL-F + AM) resulted in significant improvements compared with naive pseudo label (PL + AM), which indicates that providing more reliable and confident pseudo labels to our student network is important.
More detailed comparisons of various approaches of selecting confident pseudo label are explained in the next session.
Using AM and PL-F, our student network was more robustly trained on real world data, and now holds meaningful knowledge which can be passed back to the teacher network.
Utilizing teacher self-supervised learning (TSL) at this stage results in a notable increase in performance, compared to how basic momentum update technique performed from (PL) to (PL + MU).

\noindent{\textbf{Pseudo Label Filtering.}}
To show the effectiveness of our pseudo label filtering based on the proposed bidirectional point filtering, we experimented with well-known pseudo label filtering techniques.
The candidates were, Top k (conf), Top k (conf, class-wise), Entropy, SoftMax Max, SoftMax Avg., and ArgMax Match.
Top k filterings use the top $k$\% predictions of the pseudo label $N_\text{PL}$ based on the softmax values, where class-wise does the sorting/filtering within all instances in each class.
Entropy filtering uses the bottom $k$\% of the predictions regarding its entropy.
For SoftMax Max, SoftMax Avg., and ArgMax Match, we considered all three outputs from the teacher network, $N^\text{T}_\text{2D}$, $N^\text{T}_\text{3D}$, $N^\text{T}_\text{Fused}$.
Among the three predictions, the filtered pseudo label was generated by selecting or compositing values that were either the maximum of the softmax, the average of the softmax, or having indices match of the argmax operation.
We set $k$ to be 50, regarding the filtering approximate ratio of bidirectional point filtering.
\vspace{+0.1in}

Compared to other filtering methods, our filtering method achieved the best performance overall, as can be seen in \Tref{tab:filter_rgb}.
This suggests that our pose-aware bidirectional point filtering generates more reliable pseudo label.
This is because while other filter methods only concentrate on the predicted logit itself, while ours consider the relationship between estimated NOCS map and observed depth points.
Therefore, our approach can efficiently remove pose and depth-aware outliers.
\begin{figure*}
\begin{center}
\footnotesize
\begin{tabular}{c@{\hskip 0.005\linewidth}c@{\hskip 0.005\linewidth}c@{\hskip 0.005\linewidth}c@{\hskip 0.005\linewidth}c}
\includegraphics[width=0.19\linewidth]{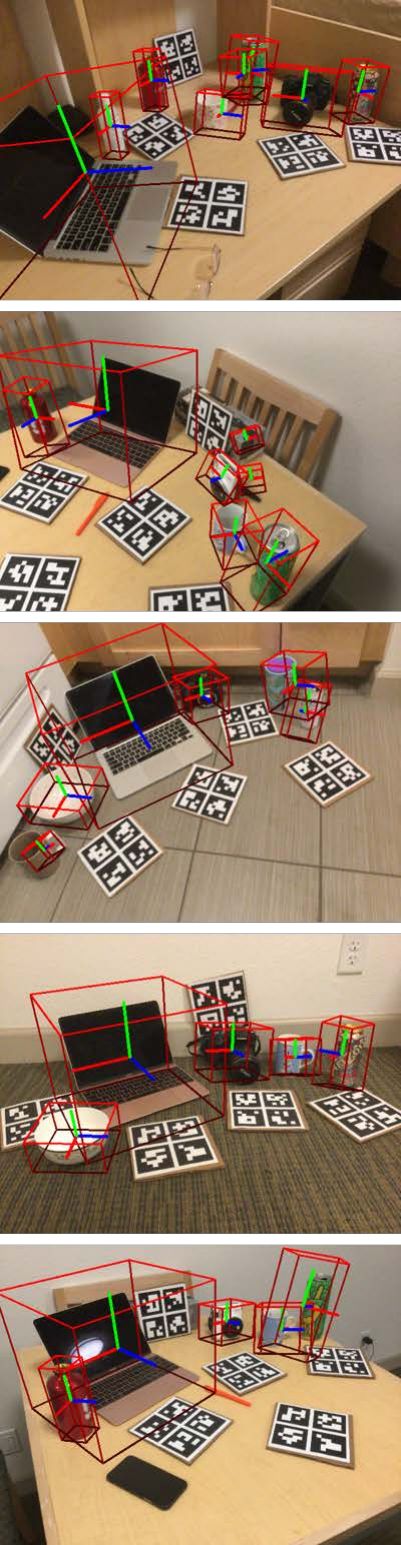} &  
\includegraphics[width=0.19\linewidth]{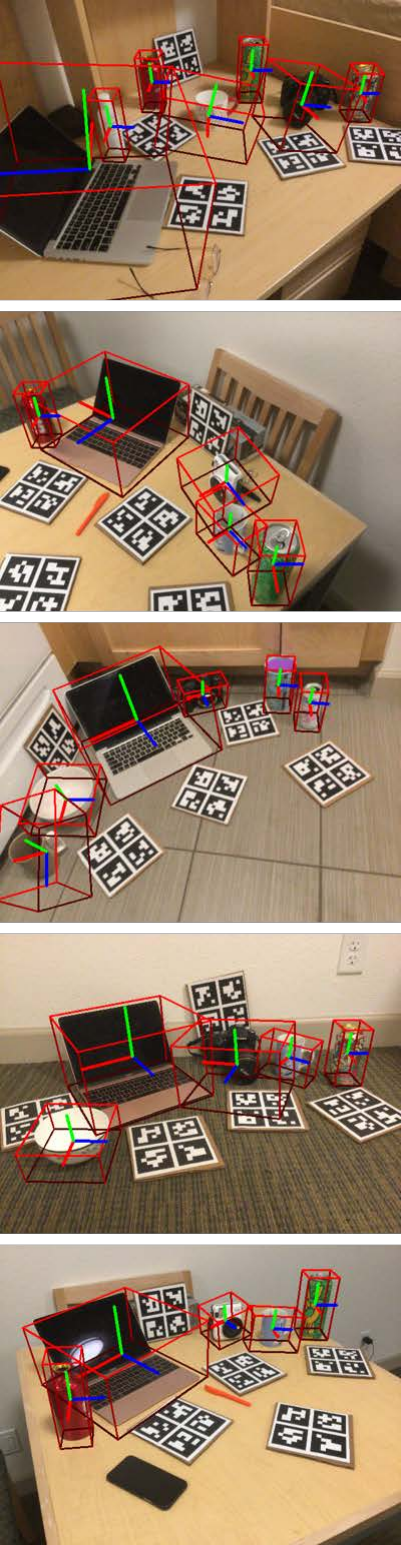} &  
\includegraphics[width=0.19\linewidth]{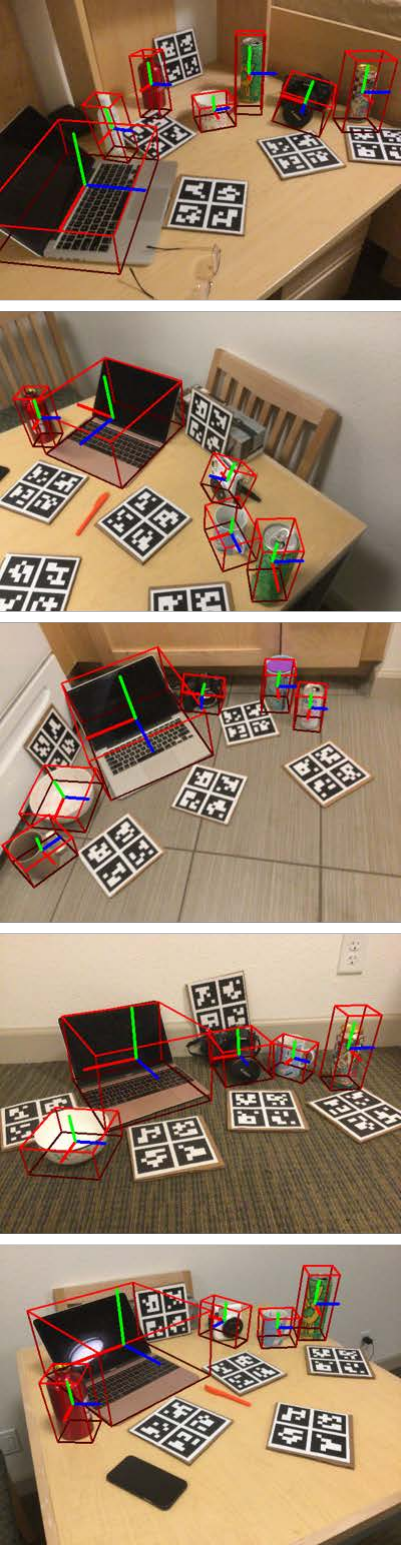} &  
\includegraphics[width=0.19\linewidth]{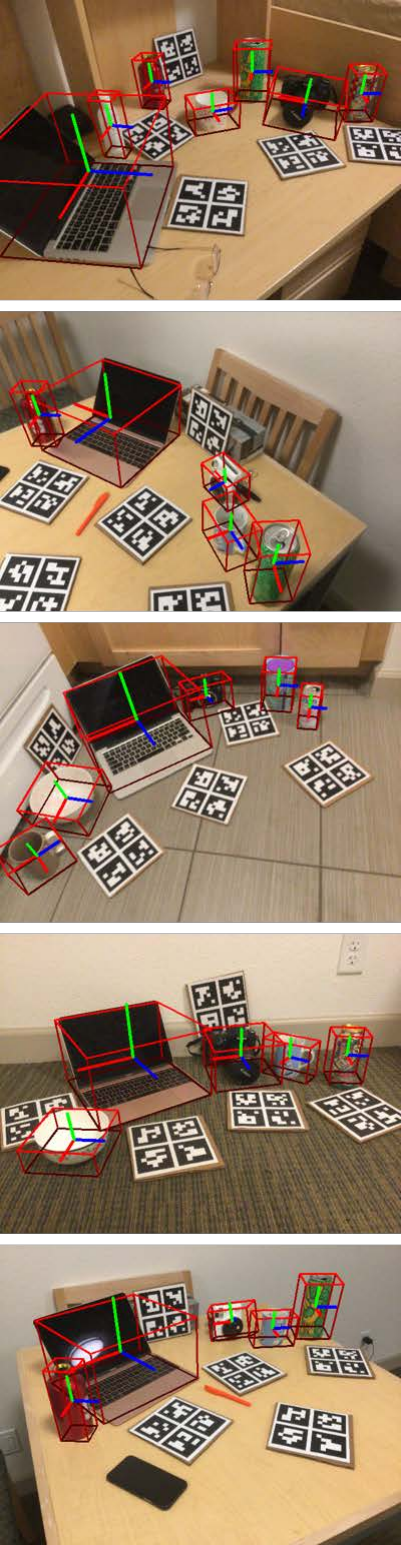} &  
\includegraphics[width=0.19\linewidth]{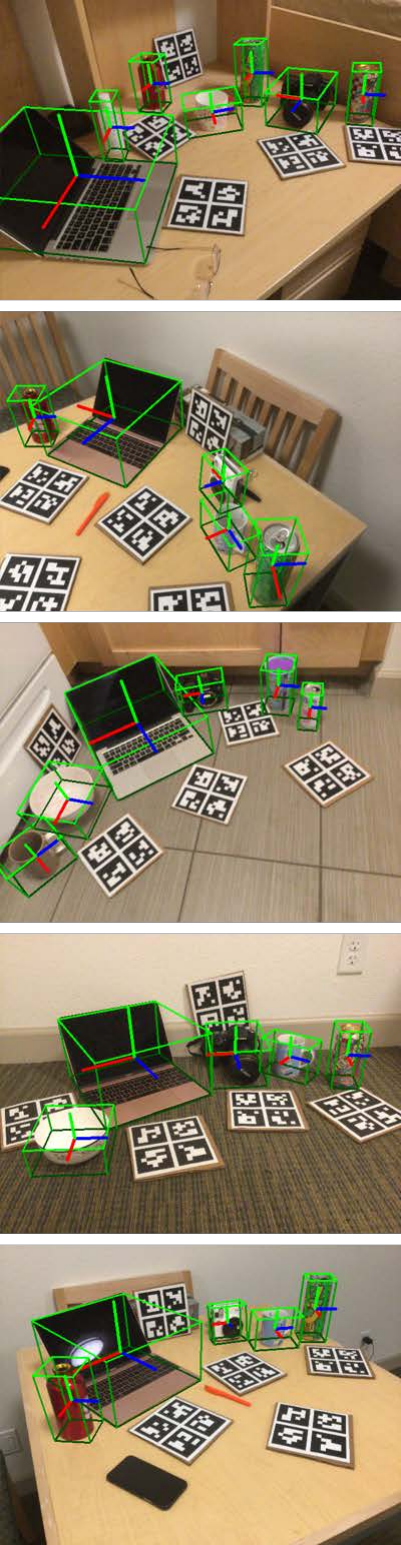} \\
{\bf (a)  NOCS~\cite{wang2019normalized}} & {\bf (b) SPD~\cite{Tian2020prior}}  & {\bf (c) DualPoseNet~\cite{lin2021dualposenet}} & {\bf (d) UDA-COPE (Ours)} & {\bf (e) Ground Truth (GT)}  \\
\end{tabular}
\end{center}
\caption{\textbf{Qualitative comparison on the REAL275 dataset.}}
\label{fig:qualitative}
\end{figure*}

\noindent{\textbf{Reliability of our Pseudo label.}}
\Fref{fig:gt_error} visualizes some examples of the real training set with GT labels and our pseudo labels.
6D poses were obtained and visualized using the Umeyama algorithm~\cite{umeyama1991least}, using GT NOCS map and our pseudo label NOCS map.
The real data annotations were mainly performed automatically using aruco markers.
For some of the failure cases, additional ICP or manual human annotations were needed.
Therefore, frames with inaccurate labels exist, which might disrupt supervised training.

However, there were cases where our method generated more accurate annotations than the GT provided by the dataset.
For example, the GT image of the first column shows that all objects have relatively wrong poses due to the occluded aruco markers in the image, while our prediction seems much more reliable.
It shows that our pseudo label is not only reliable but also sometimes more accurate than the GT, which means that our proposed approach successfully addresses both the real-world data scarcity and unreliability issues.



\subsection{Qualitative Results}
\Fref{fig:qualitative} shows qualitative results on the REAL275 dataset.
We compare our results with some of the supervised methods, NOCS~\cite{wang2019normalized}, SPD~\cite{Tian2020prior} and DualPoseNet~\cite{lin2021dualposenet}.
Our method estimated pose and sizes more accurately than NOCS and SPD, especially on cameras and laptops.
Compared to the state-of-the-art approach, DualPoseNet, ours exhibited comparable predictions, although it was not trained with the GT labels of the real dataset.

\section{Limitations and Future Work}
To the best of our knowledge, ours is the first approach that tries to solve unsupervised domain adaption for category-level 6D pose estimation using unlabeled RGB-D data.
Therefore, there may exist some issues or future directions to be addressed. For example, our pose estimation depends on object classification, detection, and segmentation to produce  appropriately  cropped  images and sampled depth points, thus sensitive to the performance of the off-the-shelf segmentation pipeline. 
Also, while the proposed method utilizes a single frame RGB-D image, we may utilize geometric constraints from the video input.

\section{Conclusions}
We propose UDA-COPE, unsupervised domain adaptation for category-level object pose estimation which addresses the real-world lack-of-label problem using multi-modality (RGB-D).
Specifically, we designed a bidirectional point filtering method to filter noisy pseudo labels, and observed depth points, where the filtered depth points improve the robustness of the teacher network, and the filtered pseudo label helps efficient student network training. Both provide for better domain adaptation with real-world pose estimation.
Experiments showed that our proposed pipeline and pose-aware point filtering results were comparable to or sometimes better than the performance of fully supervised approaches. 
\section*{Acknowledgment}
%
This work was conducted by Center for Applied Research in Artificial Intelligence(CARAI) grant funded by Defense Acquisition Program Administration(DAPA) and Agency for Defense Development(ADD) (UD190031RD).
\vfill

{
\small
\bibliographystyle{cvpr}
\bibliography{cvpr}
}

\end{document}